\documentclass[10pt,twocolumn,letterpaper]{article}

\usepackage{iccv}
\usepackage{times}
\usepackage{epsfig}
\usepackage{graphicx}
\usepackage{amsmath}
\usepackage{amssymb}
\usepackage{algorithm}\usepackage{algorithmic}

\usepackage{multirow}
\usepackage{colortbl}
\usepackage{arydshln} 
\usepackage{booktabs}

\usepackage{multirow}
\usepackage{nicematrix} 

\definecolor{citecolor}{HTML}{0071bc}
\usepackage[pagebackref=false,breaklinks=true,letterpaper=true,colorlinks,citecolor=citecolor,bookmarks=false]{hyperref}
\usepackage{pifont}
\usepackage{bm}
\usepackage{multirow}
\usepackage{bbding}
\usepackage{booktabs}
\usepackage{subfig}
\usepackage{float}
\usepackage{colortbl}

\iccvfinalcopy 

\def\iccvPaperID{5103} 

\ificcvfinal\fi

\begin{document}

\title{Dataset Condensation via Generative Model}

\author{%
   David Junhao Zhang$\textsuperscript{\rm 1}$\thanks{Work is partially done during an internship at ByteDance.},  Heng Wang$\textsuperscript{\rm 2}$, 
 Chuhui Xue$\textsuperscript{\rm 2}$, Rui Yan$\textsuperscript{\rm 2}$, Wenqing Zhang$\textsuperscript{\rm 2}$, \\ 
 Song Bai$\textsuperscript{\rm 2}$, Mike Zheng Shou$\textsuperscript{\rm 1}$\thanks{Corresponding Author.}
 \\
 \\
  $~\textsuperscript{\rm 1}$  Show Lab, National University of Singapore $~\textsuperscript{\rm 2}$  Bytedance  \\
  \\  }

\maketitle
\ificcvfinal\thispagestyle{empty}\fi


\begin{abstract}
Dataset condensation aims to condense a large dataset with a lot of training samples into a small set. Previous methods usually condense the dataset into the pixels format. However, it suffers from slow optimization speed and large number of parameters to be optimized. When increasing image resolutions and classes, the number of learnable parameters grows accordingly, prohibiting condensation methods from scaling up to large datasets with diverse classes. Moreover, the relations among condensed samples have been neglected and hence the feature distribution of condensed samples is often not diverse.  To solve these problems, we propose to condense the dataset into another format, a generative model. Such a novel format allows for the condensation of large datasets because the size of the generative model remains relatively stable as the number of classes or image resolution increases.Furthermore, an intra-class and an inter-class loss are proposed to model the relation of condensed samples. Intra-class loss aims to create more diverse samples for each class by pushing each sample away from the others of the same class. Meanwhile, inter-class loss increases the discriminability of samples by widening the gap between the centers of different classes. Extensive comparisons with state-of-the-art methods and our ablation studies confirm the effectiveness of our method and its individual component. To our best knowledge, we are the first to successfully conduct condensation on ImageNet-1k.
\end{abstract}
\begin{figure}[t]
  \centering
   \includegraphics[width=1\linewidth]{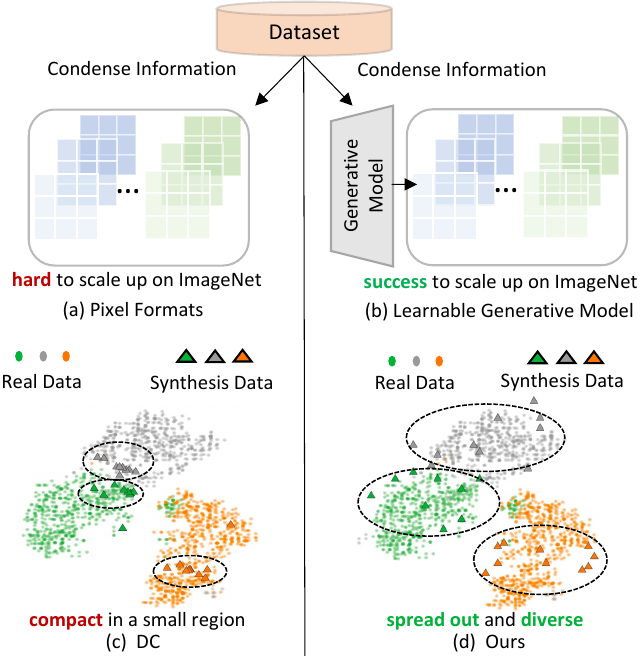}
   \caption{(a) Existing works condense information into pixels. (b) We condense information into a generative model. (c) The distribution of condensed images from DC \cite{dc}. (d) The distribution of condensed images from our method.}
   
   \label{ab}
\end{figure}

\section{Introduction}





Despite the great success of deep neural networks on a wide range of computer vision and machine learning tasks\cite{swin,cswin,pvt,t2t,timesformer,motionformer,video_swin,video_Transformer,vidtr}, it requires large-scale datasets to perform training and many runs of parameters tuning. Consequently,  computational and environmental resources turn out to be heavily consumed.

 Several works \cite{dd,citetwo,citeone,citethreeone,mtt,cafe,infobatch,dim,dream,dataset}, attempt to reduce the high training costs by mining and compressing large-scale datasets. Traditional approaches settle on a subset of the original dataset, including active learning for valuable samples labeling \cite{tong2001support} and coreset selection \cite{sener2018active,baykal2018small}. These approaches are limited by the representation capability and coverage of the selected samples. Recent studies propose a new task, namely dataset condensation (DC) or distillation (DD), to condense the large dataset into a very small number of synthetic images, while maintaining promising results of the network trained on such a small set. Starting from the seminal work \cite{dd}, various methods have been developed to improve dataset condensation, including  keeping network gradient \cite{dc}, feature \cite{cafe}, mutual information \cite{data-para} and  parameter trajectory \cite{mtt} consistent between synthetic and original sets.  All of them aim to compress the information contained in the original large dataset into a small set of synthetic images.


Despite being promising,  existing dataset condensation approaches usually face two major challenges. \textbf{(1)} First, the condensation process is slow and difficult to scale up to large datasets such as ImageNet-1K. This is because that, current methods try to distill information of original dataset directly into pixels by treating each pixel of synthetic images as a learnable parameter, whose number is proportional to the resolution and number of classes. For example, 1K classes, 128 $\times$ 128 resolutions and 10 images per class would lead to a total number of 1.5G parameters. Backpropagation on such a large number of parameters makes the optimization process extremely slow. Also, it is hard to optimize such a large amount of parameters, prohibiting dataset condensation methods from scaling up to large datasets \textit{e.g., ImageNet-1K}. \textbf{(2)} Second, since network used for data distillation is usually optimizing towards intra-class compactness \cite{wen2016discriminative}, the distribution of synthesis images in each class  tends to be clustered into a compact small region as shown in Fig.\ref{ab} (c). Such lacking of diversity precludes the synthetic images being representative enough, often leading to overfitting when training on them.


To addresse the above challenge (1), as illustrated in Fig.\ref{ab} (b), we propose to condense the information into an alternative format, namely a generative model rather than the pixels. Specifically, the generative model comprises of a codebook and a subsequent generator, both of which are optimized to contain as much information as possible from the original dataset. To synthesize an image, we initially sample a code from the learned codebook, and the generator serves as an information carrier, conveying the information from latent space to images. It takes the sampled code as input and generates an image containing dense information. Additionally, the generator is conditioned on a class embedding, which controls the synthetic
image to be class-specific. Importantly, both the codebook and generator are shared by different classes, and each code is a 1D learnable vector, resulting in fewer parameters that are less affected by the increasing number of classes and resolutions. Therefore, this innovative condensed format facilitates condensation of datasets with diverse classes and higher resolutions, such as ImageNet-1K \cite{deng2009imagenet}.

To overcome the above challenge (2), we introduce an intra-class diversity loss and an inter-class discrimination loss to improve the representation capability of the condensed dataset. Concretely, the intra-class loss regards every two samples of the same class as a negative pair, while each sample and its corresponding class center as a positive pair. In this manner, synthetic samples of the same class are spread out while are not too far away from their respective class centers. Meanwhile, the inter-class loss enlarges the  distance between samples of different classes and therefore the samples can be more easily distinguished. Thanks to these two losses, our condensed dataset exhibits a higher degree of diversity and wider coverage of information, hence is more favorable than previous condensed datasets.


We summarize our contributions as below:
\vspace{-6pt}
\begin{itemize}
  \item  Instead of directly condensing information of a large dataset into synthetic images, we propose to condense the information into a generative model, which consists of an information carrier generator and a codebook. 
  \vspace{-4pt}
 \item  Such a novel condensed format enables us to successfully scale up condensation on ImageNet-1K \cite{deng2009imagenet} for the first time.
   \vspace{-4pt}
\item We devise the intra-class diversity loss and the inter-class discrimination loss to enhance the representation capability of synthetic images and improve the generalization ability.
  \vspace{-4pt}
   \item Extensive experiments on standard benchmarks prove the superiority of our method over the existing condensation methods. The ablation studies confirm the effectiveness of each proposed component.
\end{itemize}

\section{Related Work}

\noindent\textbf{Dataset Condensation.}
Dataset condensation techniques have advanced in a number of applications including data privacy \cite{sucholutsky2020secdd,tsilivis2022robust,dong2022privacy}, neural architecture search \cite{dc,dm}, federate learning \cite{goetz2020federated,zhou2020distilled} and continue learning \cite{dc,dm,dsa}  since the seminal work \cite{dd}. Following works significantly improve the results by surrogating optimization objectives or proposing efficient optimization approaches. Worthwhile objective, e.g., trajectory matching \cite{mtt}, distribution \cite{dm} and feature alignments \cite{cafe} as well as valuable optimization adjustment e.g., synthetic-data parameterization \cite{data-para}, neural feature regression \cite{neural_regress}, soft label \cite{soft-label}, infinitely wide convolution networks \cite{infinite}, contrastive signals \cite{contrastive} and differentiable siamese augmentation \cite{dsa} are conducive to synthesizing informative images. Besides, \cite{remember,factor} propose to factorize synthetic images into the image basis and a couple of coefficients to retrieve.

However, above methods regard pixels or image basis as  learnable parameters, the number of which grows linearly with number of classes and resolution. This hinders their scalability to the dataset with diverse classes. Instead of directly optimizing  pixels/image basis, we condense information into a generative model then synthesize images. Moreover, modeling relations among synthesis images are ignored in previous approaches. On the contrary, we design an inter and inra class loss to mining relations thereby generating more informative and diverse synthesis images. For the first time to scale up condensation methods on Imagenet-1K, our work is cocurrent with \cite{cui2022scaling}.

\noindent\textbf{Coreset Selection.}
The selection of a coreset or subset is the classical method for decreasing the overall size of the training set \cite{citeone,citeseven,citefifteen,citefourone}.
The majority of these techniques involve making small, incremental selections of key data points on the basis of heuristic selection criteria. \cite{threefive} assesses the forgettability of trained samples during the network's training process and eliminates difficult-to-forget samples. \cite{citetwo} attempts to achieve maximal diversity of samples within the gradient space. \cite{citethreeone} selects data points that are in close proximity to the cluster centers for consideration. However, it cannot be ensured that the selected subset is optimal for training deep neural networks as these heuristic selection criteria were not specifically designed to function with them. Furthermore, greedy algorithms for sample selection cannot ensure that the selected subset is the most optimal to meet the desired criteria. Instead of selecting samples, we aims to synthesize a small set of samples to benefit network training.

\vspace{0.2cm}
\noindent \textbf{Generative Model.} Generative models, such as those explored in \cite{gan1,gan2,axgan,biggan,stylegan2ada,goodfellow2020generative}, have been developed for the purpose of creating realistic images, and have been applied in a range of applications, including image manipulation \cite{man1,man2}, inpainting \cite{inpaint2,inpaint1}, super-resolution \cite{super}, image-to-image translation \cite{chu2017cyclegan}, and object detection \cite{object1,object2}. On one hand, the effectiveness of these GAN-generated images for training models is comparable to that of randomly selected real images. On the other hand, some methods have utilized GAN to generate datasets, such as the work of \cite{xian2018feature} which proposes to generate features for unseen classes in zero-shot learning, and \cite{zhang2021datasetgan} which introduces a DatasetGAN for creating semantic labels. However, these methods aim to generate massive training samples and/or annotate pixel labels. Differently, IT-GAN \cite{zhao2022synthesizing} proposes to inverse information into latent codes only, but the number of synthetic training samples is not reduced and is equal to the number of samples in the original dataset. Thus, the training cost is still high.   Furthermore, the relation among generated  samples is ignored in these methods.

Apart from above methods, our method aims to condense a  dataset into a whole generative model and generate a very small set of training samples to reduce training cost. Meanwhile, we minimize our inter and intra class relation loss to ensure that the synthesized images are diverse and discriminative enough, resulting in a condensed dataset with strong representative capability. Differences in details between our method
and IT-GAN \cite{zhao2022synthesizing} can be found in Section \ref{it}. Moreover, our work is cocurrent with DiM \cite{dim}.
\begin{figure*}[t]
\begin{center}
\includegraphics[width=0.93\textwidth]{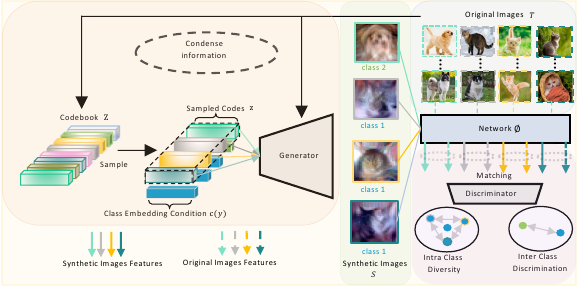}
\end{center}
\vspace{-0.5cm}
\caption{
Overall Method. We aim to condense a whole dataset into a generative model.  Given the sampled codes from the codebook and class conditions, generator synthesizes images. The generative model is optimized  to contain maximum information  by aligning synthesis images features to real image features and modeling the relations among synthesis samples.
}\label{overview}
\end{figure*}

\section{Method}

\subsection{Overview}

\begin{table}[h]
\setlength{\tabcolsep}{3pt}
    \centering
	\resizebox{1\columnwidth}{!}{
\begin{tabular}{l|cccr}
\hline
     & format     & scalable  & sample relations \\
     \hline
DD \cite{dd}  & pixels as parameters   &       	hard       &      	ignored                   \\
CAFE \cite{cafe} & pixels as parameters   &      	hard        &     ignored                     \\
MTT \cite{mtt}  & pixels as parameters    &         	hard     &    	ignored                \\
OURS  & codebooks+generator &    \checkmark          &     \checkmark                  \\ 
\hline
\end{tabular}}
\caption{Comparisons with other SOTA methods. Scalable means scalability to dataset with diverse classes.}
\label{summ}
\end{table}

Given a large dataset $\mathcal{T}=\left\{\left(x_i, y_i\right)\right\}_{i=1}^{|\mathcal{T}|}$ with large amounts of training samples $x_{i}$ and corresponding labels $y_{i}$, dataset condensation aims to get a small synthetic set $\mathcal{S}=\left\{\left(\tilde{x_i}, \tilde{y_i}\right)\right\}_{i=1}^{|\mathcal{T}|}$ whose size $|\mathcal{S}|$ is much smaller than $|\mathcal{T}|$. As shown in Fig.\ref{overview}, instead of directly condensing information into pixels, we distill information into a generative model. The class embeddings are sent into the generator together with sampled codes from the codebook. They serve as a detonator to boom the information squeezed in the  generator and enable it synthesizing informative images. Meanwhile, the relation of synthetic images is constrained with our inter and intra-class loss. The main differences with other SOTA methods are summarized in Tab.\ref{summ}.

\subsection{Condense Dataset into Generative Model}

\noindent\textbf{Codebook.} Instead of treating condensed images as learnable parameters and directly optimizing them, we propose  a learnable codebook $Z\in \mathbb{R}^{K\times C}$, where $K$ is the number of images per class and $C$ is the latent dimension. In this way, the number of learnable parameters is no longer significantly effected by the number of classes and spatial resolutions. This makes it possible to condense dataset that has a large number of diverse classes. Since different classes share the same codebook, the learned codebook contains valuable mutual information across classes and has strong representative capability.
\vspace{-0.02cm}

\noindent\textbf{Generative Models.} 
Given sampled information squeezed codes $z$ from $Z$,  our generator is optimized to  comprise maximum information and acts as an information carrier, conveying the information from the latent space to the images. The primary goal of the generator is to synthesize informative images, rather than realistic ones. As indicated by \cite{dd}, the generated images are usually out of the distribution of original datasets. Therefore, training on these images would lead to inferior performance. To alleviate this effect, we take the feature embeddings instead of images as input into discriminator, which enables the feature distribution of synthetic images to align or cover the distribution of original datasets better. Thus we arrive at an adversarial loss as follows,

\begin{equation}
\begin{aligned}
\min _G \max _D \mathcal{L}_{adv}=& E[\log D(\phi_{\theta}(x), c(y))]+\\
& E[\log (1-D(\phi_{\theta}(\tilde{x}), c(y)))],
\end{aligned}
\end{equation}
where  $D$ is a discriminator, $\tilde{x}$=$G(z, c(y))$, c(y) is a class condition, $G$ is a generator and $\phi_{\theta}()$ represents a feature extractor on a task specific dataset. 


Furthermore, we add a classifier head on top of the features from both real and synthetic images inspired by \cite{axgan}. Updating the generator and codebooks via this classification loss $\mathcal{L}_c$ enables the synthesis images to share the same category-based semantics as real images. The overall GAN loss can be denoted as $\mathcal{L}_{GAN}=\mathcal{L}_{adv}+\mathcal{L}_{c}$. 
\vspace{-0.05cm}

\noindent\textbf{Class Embedding.} The class embedding plays a role of a fuse to enable generator to boom the condensed information. Hence, we propose to condition on more informative class embeddings instead of one-hot labels.  Specifically, we extract and spatially
pool the lowest resolution features of all images in dataset using a well-trained feature extractor and calculate the mean per class in the dataset.
Then a projection layer is applied to make it balance with $z$ code and we
concatenate them together as the input for generator.

\vspace{-0.05cm}
\noindent\textbf{Matching Loss.} To make synthetic images more informative and representative, we simply adopt the feature matching loss \cite{cafe} as our training objective to update our generator and  codebook. For each synthetic image $\tilde{x}$, since the number of original images that belong to the same class is large, we  sample a subset of original images as a association. As shown in Fig.\ref{overview}, we pass the synthetic image and its association respectively through a network denoted as $\phi_{\theta}$. Then the feature ${\boldsymbol{f}}_{ l}^{\mathcal{S}}$ of $\tilde{x}$ and the mean feature $\overline{\boldsymbol{f}}_{l}^{\mathcal{T}}$ of  its
 association from every layer $l$ are obtained. In the end, the MSE is applied to calculate the feature distribution
matching loss $\mathcal{L}_{\mathrm{f}}$ between the synthetic image and  its association(original images) for every layer. The loss is formulated as 
\begin{equation}
\mathcal{L}_{\mathrm{f}}= \sum_{l=1}^L\left|{\boldsymbol{f}}_{ l}^{\mathcal{S}}-\overline{\boldsymbol{f}}_{l}^{\mathcal{T}}\right|^2.
\end{equation}

\subsection{Intra-Class Diversity and Inter-Class Discrimination.}
\noindent\textbf{Intra-Class Diversity.} As mentioned in the introduction, relations among synthetic images are neglected and therefore, as shown in Fig.\ref{ab} (c), the synthetic images of each class have pretty low diversity. Since the synthetic images for the same class are not diverse enough, training on them can easily result in overfitting. To overcome this issue, we treat any two samples of the same class as a negative pair, pushing their feature embedding away from each other. However, this alone may widen the sample distribution of one class too much, and hence the sample distributions of different classes are overlapped. To remedy this effect, we aim to constrain the scope of distribution of each class. Specifically, we introduce class center for each class; each sample and its corresponding class center form a positive pair, and their features are to be pulled closer. With these two kinds of pairs, we can achieve a balance that features of the same class samples are diverse enough while can still be distinguished against features of other classes. Formally, we design an intra-class diversity loss:
 
\begin{equation}
\mathcal{L}_{intra}=
-\log \frac{ e\left(\left\langle \boldsymbol{f}^{\mathcal{S}}, c(y)\right\rangle \right)}{\sum_{i=1}^{K-1} e \left(\left\langle \boldsymbol{f}^{\mathcal{S}}, \boldsymbol{f_i}^{\mathcal{S}_{y}}\right\rangle \right)+e \left(\left\langle \boldsymbol{f}^{\mathcal{S}}, c(y)\right\rangle \right)},
\end{equation}
where $e$ is $\exp\left(\frac{.}{\tau}\right)$, $\left\langle\right\rangle$ is the dot product and $\boldsymbol{f}^{\mathcal{S}}$ is the feature of the synthetic image $\tilde{x}$ obtained from the last layer of feature extraction network $\phi_{\theta}$. $c(y)$ is a class embedding and $\boldsymbol{f_i}^{\mathcal{S}_{y}}$ is the feature of another sample, whose class is same as  $\tilde{x}$. 

\noindent\textbf{Inter-Class Discrimination.} To further make our synthetic samples of different classes  distinguishable from each other, we aim to enlarge the margin between feature distributions of different classes. Inspired by \cite{yu2019single}, we propose an inter-class discrimination loss:
\begin{equation}
\mathcal{L}_{inter}= \sum_{\substack{c_A=1 \\ c_A \neq c_B}}^C \sum_{c_B=1}^C \max \left(\tau_{\mathrm{m}}-\left\|\overline{\boldsymbol{f}}^{\mathcal{S}_{c_A}}-\overline{\boldsymbol{f}}^{\mathcal{S}_{c_B}}\right\|, 0\right),
\end{equation}
where $\overline{\boldsymbol{f}}^{\mathcal{S}_{c_B}}$ and $\overline{\boldsymbol{f}}^{\mathcal{S}_{c_A}}$ are the mean values of features of different class in each synthetic images batch. When the distance between two class centers is large than margin $\tau_{m}$, the penalty is zero. This loss forces the mean center of each class to be far away from other classes and hence makes the samples of different classes more discriminative. 

\subsection{Optimization Framework}
We adopt the bi-level optimization following \cite{cafe,dc,dd,remember}. The optimization framework is illustrated in Algorithm \ref{alg:matching}. In the outer-loop, the overall condensation loss $\mathcal{L}_{con}=\mathcal{L}_{GAN}+\mathcal{L}_{f}+\mathcal{L}_{intra}+\mathcal{L}_{inter}$ is used to update the generator $G$ and codebook $Z$. Meanwhile, the network $\phi_{\theta}$ for feature matching is updated by general classification loss $\mathcal{L}_{cls}$. Unlike previous methods \cite{dc,cafe}, the $\phi_{\theta}$ is dependent on real images $|T|$ rather than synthetic images $|S|$ for training. Hence, it can force the layer statistics e.g., mean and variance produced by our synthetic images to align with those of the real data during feature matching. Moreover, it further forces the distribution of synthetic images to align with real images so that fool the discriminator.

\begin{algorithm}[!t]
    \caption{Optimization Framework}
    \label{alg:matching}
\begin{algorithmic}
\STATE {\bfseries Input:} Training data $\mathcal{T}$
\STATE {\bfseries Notation:} Generator $G$, codebooks $Z$, Synthetic dataset $\mathcal{S}$, Network for feature matching  $\phi_{\theta}$ and hyperparameters $\lambda$, $\alpha$ are learning rates.\\
\REPEAT 
\STATE Initialize network parameter $\theta_1$
\FOR{$n=1$ {\bfseries to} $N$}
    \FOR{$i=1$ {\bfseries to} $M$}
    \STATE Sample  multiple $z$ codes from $|Z|$
    \STATE Generate the synthetic images set $\mathcal{S}_i$ via $G$ 
    \STATE Sample their associations $\mathcal{T}_i$ from $\mathcal{T}$ \\
    \STATE Compute $\mathcal{L}_{con}$ via $\mathcal{T}_i$ and $\mathcal{S}_i$ using $\phi_{\theta}$ \\
    \STATE Update $(Z,G)\leftarrow(Z,G)-\alpha\nabla_{{(Z,G)}}\mathcal{L}_{con}$
    \ENDFOR
    
    \STATE Inner loop update $\theta\leftarrow\theta-_{\nabla}\mathcal{L}_{cls}(\mathcal{T})$
\ENDFOR

\UNTIL{Converge}

\end{algorithmic}
\end{algorithm}

\begin{table*}[tp]
\renewcommand\arraystretch{1.0}
\centering

\setlength{\tabcolsep}{2pt}

 \centering
	\resizebox{1.0\textwidth}{!}{
\begin{tabular}{cccccccccccccccc}
\hline
\hline
\multirow{2}{*}{}           & \multirow{2}{*}{IPC} & \multirow{2}{*}{Ratio \%} & \multicolumn{4}{c}{Coreset Selection} 
&\multicolumn{6}{c}{Condensation}     &&\multirow{2}{*}{Whole Dataset} \\ 
                            &                          &                           & Random          & Herding         & K-Center       & Forgetting &DD$^{\dag}$ \cite{dd}&LD$^{\dag}$   &DC \cite{dc} &DSA \cite{dsa} &CAFE \cite{cafe}        & MTT \cite{mtt}   &OURS   \\ \midrule
\multirow{3}{*}{MNIST}          & 1   & 0.017  & 64.9$\pm$3.5  & 89.2$\pm$1.6  & 89.3$\pm$1.5  & 35.5$\pm$5.6    &-&60.9$\pm$3.2& 91.7$\pm$0.5 &88.7$\pm$0.6 &93.1$\pm$0.3   &-- & \textbf{93.7$\pm$0.9}& \multirow{3}{*}{99.6$\pm$0.0} \\ 
                                & 10  & 0.17   & 95.1$\pm$0.9  & 93.7$\pm$0.3  & 86.4$\pm$1.7  & 68.1$\pm$3.3   &79.5$\pm$8.1&87.3$\pm$0.7& 97.4$\pm$0.2 &97.8$\pm$0.1 &97.5$\pm$0.1 &-- &\textbf{98.4$\pm$0.1} &\\ 
                                & 50  & 0.83   & 97.9$\pm$0.2  & 94.8$\pm$0.2  & 97.4$\pm$0.3  & 88.2$\pm$1.2   &-&93.3$\pm$0.3& 98.8$\pm$0.2 &\textbf{99.2$\pm$0.1} &98.9$\pm0.2$ &-- &\textbf{99.2$\pm$0.1}& \\ \midrule

\multirow{3}{*}{FashionMNIST}   & 1   & 0.017  & 51.4$\pm$3.8  & 67.0$\pm$1.9  & 66.9$\pm$1.8  & 42.0$\pm$5.5  &-&- & 70.5$\pm$0.6 &70.6$\pm$0.6 &77.1$\pm$0.9&-- & \textbf{79.2$\pm$1.0}& \multirow{3}{*}{93.5$\pm$0.1} \\ 
                                & 10  & 0.17   & 73.8$\pm$0.7  & 71.1$\pm$0.7  & 54.7$\pm$1.5  & 53.9$\pm$2.0    &-&-& 82.3$\pm$0.4 &84.6$\pm$0.3 &{83.0$\pm$0.4} &-- &\textbf{87.3$\pm$0.4} &         \\ 
                                & 50  & 0.83   & 82.5$\pm$0.7  & 71.9$\pm$0.8  & 68.3$\pm$0.8  & 55.0$\pm$1.1  &-&- & 83.6$\pm$0.4 &88.7$\pm$0.2 &{88.2$\pm$0.3} &-- &\textbf{88.8$\pm$0.3} &\\ \midrule

\multirow{3}{*}{SVHN}           & 1   & 0.014  & 14.6$\pm$1.6  & 20.9$\pm$1.3  & 21.0$\pm$1.5  & 12.1$\pm$1.7  &-&-&  31.2$\pm$1.4 &27.5$\pm$1.4 &42.9$\pm$3.0 &-- &\bf{51.8$\pm$2.2} &\multirow{3}{*}{95.4$\pm$0.1} \\ 
                                & 10  & 0.14   & 35.1$\pm$4.1  & 50.5$\pm$3.3  & 14.0$\pm$1.3  & 16.8$\pm$1.2   &-&-& 76.1$\pm$0.6 &79.2$\pm$0.5 &77.9$\pm$0.6 & -- &\textbf{82.1$\pm$0.5}   &   \\
                                & 50  & 0.7    & 70.9$\pm$0.9  & 72.6$\pm$0.8  & 20.1$\pm$1.4  & 27.2$\pm$1.5   &-&-& 82.3$\pm$0.3 &\textbf{84.4$\pm$0.4} &{82.3$\pm$0.4} &-- &\textbf{84.4$\pm$0.5} &\\ \midrule 

\multirow{3}{*}{CIFAR10}        & 1   & 0.02   & 14.4$\pm$2.0  & 21.5$\pm$1.2  & 21.5$\pm$1.3  & 13.5$\pm$1.2   &-&25.7$\pm$0.7& 28.3$\pm$0.5 &28.8$\pm$0.7 &{31.6$\pm$0.8} &46.3$\pm$0.8 &\textbf{48.2$\pm$0.8} & \multirow{3}{*}{84.8$\pm$0.1}         \\ 
                                & 10  & 0.2    & 26.0$\pm$1.2  & 31.6$\pm$0.7  & 14.7$\pm$0.9  & 23.3$\pm$1.0   &36.8$\pm$1.2&38.3$\pm$0.4& 43.4$\pm$0.5 &52.1$\pm$0.5 &{50.9$\pm$0.5}  &65.3$\pm$0.7  & \textbf{66.2$\pm$0.5} &       \\ 
                                & 50  & 1      & 43.4$\pm$1.0  & 40.4$\pm$0.6  & 27.0$\pm$1.4  & 23.3$\pm$1.1   &-&42.5$\pm$0.4& 53.9$\pm$0.5 &60.6$\pm$0.5 &{62.3$\pm$0.4} &72.8$\pm$0.2 &\textbf{73.8$\pm$0.1}  &\\ \midrule
\multirow{3}{*}{CIFAR100}   & 1   & 0.2    &  4.2$\pm$0.3  &  8.4$\pm$0.3  &  8.3$\pm$0.3  &  4.5$\pm$0.3  &-&11.5$\pm$0.4& 12.8$\pm$0.3  & 13.9$\pm$0.3 &{14.0$\pm$0.3} &24.3$\pm$0.3 & \textbf{26.1$\pm$0.3}&\multirow{3}{*}{56.17$\pm$0.3}          \\
                            & 10  & 2      & 14.6$\pm$0.5  & 17.3$\pm$0.3  &  7.1$\pm$0.2  &  9.8$\pm$0.2  &-&-& 25.2$\pm$0.3  & 32.3$\pm$0.3 &{31.5$\pm$0.2} &40.1$\pm$0.4     &  \textbf{41.9$\pm$0.3}& \\
                             & 50  & 10      & 30.0$\pm$0.4  & 33.7$\pm$0.5  &  30.5$\pm$0.3  &  -  &-&-& -  & 42.8$\pm$0.4 & {42.9$\pm$0.2} &47.7$\pm$0.2      &\textbf{48.5$\pm$0.2} & \\
                            \hline
                            \hline
\end{tabular}}
\caption{The performance (testing acc. \%) comparison to state-of-the-art methods.  LD$^{\dag}$ and DD$^{\dag}$ use LeNet for MNIST and AlexNet for CIFAR10, while the rest use ConvNet for training and testing. IPC: Images Per Class, Ratio~(\%): the ratio of condensed images to whole training set. Comparisons under store memory budget are shown in Section A of supplement.}
\label{tab:sota_coreset}
\end{table*}

\begin{table}[t]
  \centering
	\resizebox{1\columnwidth}{!}{
\begin{tabular}{l|ccc|c|c}
\hline
\hline
 IPC  & Random & FrePo & DM  & Ours & Whole Dataset         \\
   \hline
1  & 0.6  $\pm$0.1  & 7.5$\pm$0.3  & 1.5$\pm$0.1& 7.9$\pm$0.5  & \multirow{4}{*}{34.2$\pm$0.4} \\
2  & 0.9 $\pm$0.1   & 9.7$\pm$0.2  & 1.7$\pm$0.1 & 10.0$\pm$0.6 &                       \\
10 & 3.8  $\pm$0.1  &  --     &--     & 17.6$\pm$1.7 &                       \\
50 & 15.4  $\pm$1.6 &   --    &  --   & \textbf{27.2 $\pm$1.6}&           \\           
\hline
\hline
\end{tabular}}
\caption{ImageNet-1K results. 1/2/10/50 images per class settings are reported.}
\label{imagenet}
\end{table}
\
\begin{table}[t]
  \centering
	\resizebox{1\columnwidth}{!}{
\begin{tabular}{l|llll}
\hline
\hline
          IPC  & 1    & 2   & 10  & 50   \\
            \hline
BigGAN \cite{biggan}     & 0.5$\pm$0.1  & 0.9$\pm$0.1 & 3.7$\pm$0.1 & 14.3$\pm$0.9 \\
VQGAN  \cite{esser2021taming}     & 0.8$\pm$0.1  & 0.9$\pm$0.1 & 3.8$\pm$0.2 & 15.7$\pm$0.8 \\
StyleGAN-XL \cite{xl}& 0.8$\pm$0.2  & 0.9$\pm$0.1 & 3.7$\pm$0.1 & 15.9$\pm$1.0 \\
\hline
Ours         & 7.9$\pm$0.5 &10.0$\pm$0.6     & 17.6$\pm$1.7    &  \textbf{27.2 $\pm$1.6} \\
\hline
\hline
\end{tabular}}
\caption{Use the synthesis images from different generative model for ImageNet-1K training. All images are resize into $64\times 64$ for fair comparisons.}
\label{ganhh}
\vspace{-0.02cm}
\end{table}

\section{Experiments}
\subsection{Dataset}
\noindent\textbf{MNIST} \cite{lecun1998mnist} contains 60000 training images and 10000 test images  with $28 \times 28$ size for handwritten digit recognition.

\noindent\textbf{Fashion MINIST} \cite{xiao2017fashion} has a training set of 60,000 samples and a test set of 10,000 examples, It is a dataset of Zalando article photos. Each example consists of a $28 \times 28$ image paired with a label from one of ten classes.

\noindent\textbf{SVHN} \cite{sermanet2012convolutional} is a real-world picture dataset to create machine learning and object recognition algorithms. It has more than 600,000 digitized photos drawn from actual data. All images are cropped into $32 \times 32$ size.

\noindent\textbf{CIFAR10/100} \cite{cifar10} comprise  $32\times32$ small colored natural images from 10 and 100 classes, respectively. Each dataset has 50,000 and 10,000 images for training and testing.

\noindent\textbf{ImageNet-1K} \cite{deng2009imagenet} spans 1000 object classes and contains 1,281,167 training images, 50,000 validation images. The version of $64\times64$ is what we use.

\subsection{Implementation Details}  We optimize  $Z$ codebook and generator using three/four blocks Convolutional Network(ConvNet). Each block has a convolution, an instance norm, a relu and a pooling layer. The generator is pretrained on task-specific datasets and consists of three blocks for 32 image size datasets or four blocks for 64 size dataset. Each block has two activation-convolution-bn layers and an upsampling layer. The discriminator is a multi-layer perceptron with a sigmoid function as the last layer. As indicated in \cite{neural_regress},  hard one-hot labels do not work well on dataset with diverse classes. This is because that,  the classes are not completely mutually exclusive. Therefore, we use the soft label extracted from a pre-trained ConvNet for classification in GAN loss to update $Z$ and $G$, when conducting experiments in Imagenet-1K dataset only.  The learning rate for updating $Z$, $G$, and $\phi_{\theta}$ is 0.01, 0.001, 0.01 with the SGD optimizer and a linear epoch decay, respectively. Following \cite{dc, cafe,mtt}, for the evaluation protocol, each model is evaluated
on 20 randomly initialized models, trained for 300 epochs on a synthetic dataset and reported with mean and standard deviation. All experiments are conducted on one A100 GPU.

\subsection{Comparisons with State-Of-The-Art}
We compare our method with other approaches in Tab.\ref{tab:sota_coreset}. The comparisons follow the standard 1/5/50 images per class (IPC)  evaluation setting. The ConvNet is utilized during the training and testing stage. Overall, the table can be separated into two groups. One is classic core selection, including Random, K-Center, and Herding selection, and the other one is recent dataset condensation methods, including LD \cite{bohdal2020flexible}, DC \cite{dc}, CAFE \cite{cafe}, MTT \cite{mtt},  and DSA \cite{dsa}.

Since \cite{remember,data-para,liu2022dataset} use the memory story budget  instead of standard IPC as the target, which leads to more images per class, we do not include them in Tab.\ref{tab:sota_coreset} for fair comparisons.  
Compared with the coreset selection method, our method surpasses them by a large margin.  Compared with the SOTA condensation approach MTT \cite{mtt}, our method achieves 1.9 \%  test accuracy improvement when training with one condensed image per class on CIFAR 10 dataset. With more condensed images (10 and 50) per class, our method consistently beat the performance of MTT. Moreover, with more complex labels on CIFRAR-100, our method also obtains outstanding results with different numbers of condensed images per class. All above performance demonstrates the superiority of our format, which condenses  information into a generative model, over the format that defines synthesis images as learnable parameters.

\subsection{ImageNet-1K Results}
It is difficult for other methods to be applied on ImageNet-1K \cite{deng2009imagenet}. This is because that, heavy number of parameters in synthesis images which is proportional to the number of classes and resolutions is hard to be optimized, causing high GPU memory. 
Instead, our generative model successfully scales up the dataset condensation methods into ImageNet-1K with diverse classes and color spaces. Concretely, as shown in Tab.\ref{imagenet}, our method can achieve 17.6$\%$ accuracy and 27.2 $\%$ accuracy with only 10 and 50 condensed images per class for training. Given that the upper bound 34.2 $\%$ for training the whole dataset, our method achieves the promising result training on such less training samples. These results further validate that condensing the information into a generative model instead of the original pixel space is meaningful and effective.

\subsection{Comparisons with Other Generative Models}
\label{it}
\textbf{Generative model optimised for high-fidelity.} 
We also utilize the synthetic images generated by different generative models including BigGAN \cite{biggan}, VQGAN \cite{esser2021taming}, StyleGAN-XL \cite{xl} for training. All images are resized to $64\times64$. As shown in Tab.\ref{ganhh},  although above generative models can synthesis high quality images, the results training on these synthetic images are nearly  same as random selections.  On the contrary, our method can synthesize more informative samples for training. The result shows that information is successfully condensed into our generative model.

\noindent\textbf{IT-GAN \cite{zhao2022synthesizing}.} Our approach differs from IT-GAN in several ways. \textbf{(1) Motivation.} IT-GAN aims to explore  if a fixed gan can generate the informative images without changing the size of dataset and reducing training cost. Our method aims to condense a dataset into a generative model and synthesize a very small size dataset to reduce training cost. \textbf{(2) Latent input.} In IT-GAN, given a fixed pre-trained generator $G$, it learns the whole latent set $Z \in \mathbb{R}^{\mathcal{T}\times d_z}$. Each latent vector $z \in \mathbb{R}^{d_z}$ corresponds to a real image $x$   in the original dataset $\mathcal{T}= \{x_i, y_i\}^{|\mathcal{T}|}_{i=1}$  and  generate one synthesis image. The number of synthesis images is not reduced and equal to the original number of images in  datasets.  In contrast,  our codebook $Z\in \mathbb{R}^{K\times d_z}$ is shared by different class, where K is the condensed number of images per class with $K << |T|$.  \textbf{(3) Condition.}   IT-GAN is an unconditional GAN and relies only on the z code, whereas our method is conditioned on the class embedding, which controls the class of the synthesized image.  Above two differences bring the feasibility  of  our method on large scale IN1K . Note that $|T|$ $=$ 1.2 million on IN1K is huge so it is hard to apply IT-GAN on IN1K. 
\textbf{(4) Generator.} The generator of IT-GAN is fixed but our generator is optimized to contain more information. \textbf{(5) Results and loss.} If we follow IT-GAN to fix generator and exclude our proposed inter and intra-class loss, the accuracy under 10-IPC on CIRAR-10 drops from 66.2\% to 57.6\%, which shows the superiority and novelty of our method. 
\begin{table}[]
\centering

\resizebox{1\columnwidth}{!}{
\begin{tabular}{ccccccc}
\hline
\hline
&   \texttt{C}\textbackslash \texttt{T} & ConvNet      & AlexNet      & VGG11          & ResNet18  &MLP           \\ 	\midrule
DC                 & ConvNet           & 53.9$\pm$0.5 & 28.77$\pm$0.7 & 38.76$\pm$1.1 & 20.85$\pm$1.0  &28.71$\pm$0.7 \\
\multirow{1}{*}{CAFE} & ConvNet           & {55.5$\pm$0.4} & {34.0$\pm$0.6} & {40.6$\pm$0.8} & 25.3$\pm$0.9 & {36.7$\pm$0.6} \\

Our &ConvNet &\textbf{73.8$\pm$ 0.3} &\textbf{53.9$\pm$ 0.4}&\textbf{60.1$\pm$ 0.2} & \textbf{54.9$\pm$ 0.7}&\textbf{51.1$\pm$ 0.5}\\
  \hline
  \hline
\end{tabular}}

\caption{The generalization ability (\%) on unseen architectures. C means the network used for condensing and T means netwoks used for testing. }
\label{cross}
\end{table}
\begin{table}[t]
    
    \centering
    \begin{minipage}[t]{0.44\linewidth}

        \centering
        \setlength{\tabcolsep}{1.0pt}
        \resizebox{\linewidth}{!}{
        
        \begin{tabular}{l|lll}
\hline
\hline
               & C-10 & C-100 & IN-Sub \\
               \hline
one-hot        & 62.2    &38.1&     20.5  \\
online feature & 64.0    &     39.9    &  22.1      \\
class feature  & 66.2    &    41.9      &  26.6  \\   \hline\hline
\end{tabular}
    }
     \caption{Ablation study on different conditions.}
  \label{tab:class}
    \end{minipage}
    \begin{minipage}[t]{0.54\linewidth}
     
        \centering
        \setlength{\tabcolsep}{1.0pt}
        \resizebox{\linewidth}{!}{
         \begin{tabular}{l|lll}
\hline\hline

                                                            & C-10 & C-100 & IN-Sub \\
                                  \hline
uniform sample                                          & 55.3    &      27.1    & 12.2  \\
\hline
random with threshold                                                       & 55.9    &   28.4       & 12.5   \\
\hline

          learnable codebooks                          & 66.2    &     41.9     & 26.6 \\
                                 \hline\hline
\end{tabular}
        }
   \caption{Ablation study on different input samples.}
\label{code}
    \end{minipage}

 \vspace{-0.5cm}
  
\end{table}

\begin{table}[]
\setlength{\tabcolsep}{2pt}
    \centering
	\resizebox{0.7\columnwidth}{!}{
\begin{tabular}{llll|lll}
\hline\hline
l\_real & l\_feature & l\_intra & l\_inter & CIFAR-10 & CIFAR-100 & IN-SUB \\
\hline
      \checkmark  &            &          &          &    63.2       &      37.2  &   20.1     \\
        &      \checkmark      &          &          & 64.7    & 39.1         & 23.4   \\
        &       \checkmark     &  \checkmark        &          & 64.9    &     39.4    & 23.5   \\
        &         \checkmark   &          &\checkmark          & 65.4    &     40.8    & 25.3   \\
        &    \checkmark        &  \checkmark        &  \checkmark        & 66.2    & 41.9     & 26.6  \\
        \hline\hline
\end{tabular}}
\caption{Ablation study of loss.}
\label{loss}
\vspace{-0.5cm}
\end{table}

\subsection{Ablation Study}

In this section, we ablate the effect of variations for proposed components including  codebook, class embedding, intra-class diversity, and inter-class discrimination loss in our method. All experiments are conducted with 10 images per class setting on three datasets including CIFAR-10/100 and ImageNet. For ImageNet, we select 100 classes as a subset from the original 1000 classes. The performance trend of different architecture on these 100 classes is similar to the original dataset.

\noindent\textbf{Generalization ability.} 
Since the condensed images are generated with ConvNet feature matching, we explore the generalization ability of our condensed images with unseen architectures, including AlexNet \cite{alexnet}, VGG11 \cite{vgg}, ResNet18 \cite{resnet} and MLP. As shown in Tab.\ref{cross}, our method achieves outstanding generalization performance on different architectures. This could be attributed to that our inter and intra-class loss improve the diversity of synthesis images for more robust generalization.

\noindent\textbf{Different class embedding}. We evaluate three different formats of class embedding as a detonator to boom the information. One-hot is a regular label for each class. Online features represent that using the feature of images extracted by network $\phi_{\theta}$ at the last step so it varies during training. The class feature is the mean feature of each class as introduced in the method. As shown in Tab.\ref{tab:class}, the class feature achieves the best results since it contains more information acquired from all images in each class. It provides strong prior emerged from a specific class  and enables the generator to synthesize informative images. Such a prior brings more improvement on the complex dataset-ImageNet, which has mutual information across classes.

\noindent\textbf{Codebook.} We demonstrate the representative capability of $|Z|$ codebook in Tab.\ref{code}. Uniform sampling means that a couple of $Z$ codes are sampled uniformly in a gaussian distribution. Random with threshold represents using truncation trick as \cite{biggan}. Compared with other two sample methods, the learnable codebook brings $10.9\%, 14.8\%, 14.4\%$ and $10.3\%, 13.5\%, 14.1\%$ on three datasets respectively. These improvements indicate the effectiveness of condensing the information into our learnable codebook.

\begin{figure*}[t]

\centering
\subfloat[Original CIFAR10 images.]{\includegraphics[width = .32\linewidth]{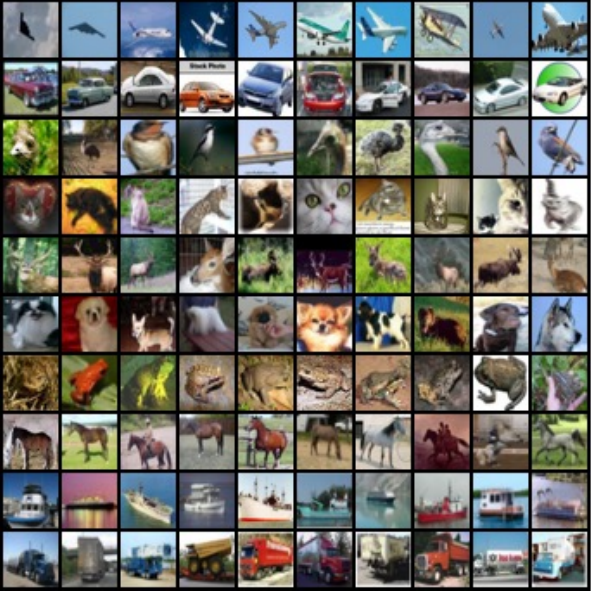}\label{fig:vis_a}}\hfill
\subfloat[The synthetic images of CAFE.]{\includegraphics[width = .32\linewidth]{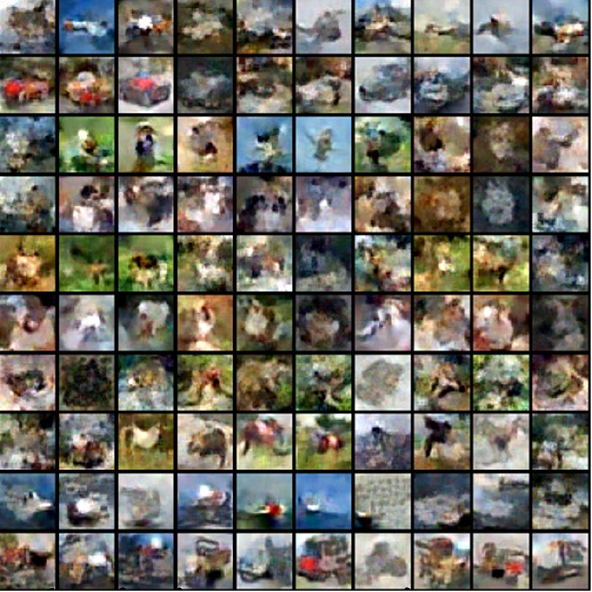}\label{fig:vis_b}}\hfill
\subfloat[The synthetic images of ours.]{\includegraphics[width = .31\linewidth]{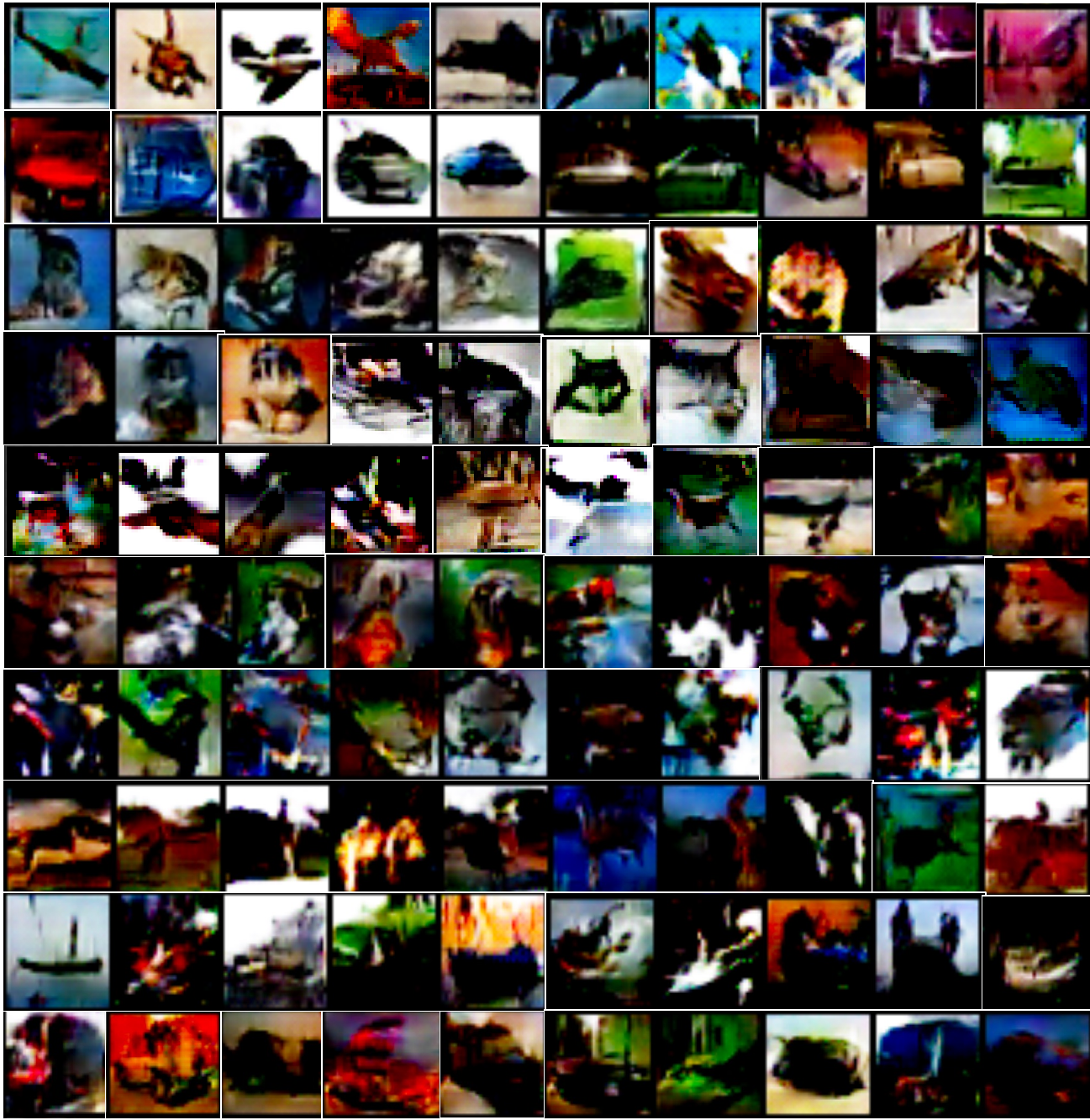}\label{fig:vis_c}}
\caption{Visualizations of original images in CIFAR10, and the synthetic images generated by CAFE and our method. CAFE  is initialized from random noise. The condensation ratio is 10 images per class.}     
\label{vis}
\end{figure*}

\begin{figure*}[h]
\begin{center}

\includegraphics[width=1\textwidth]{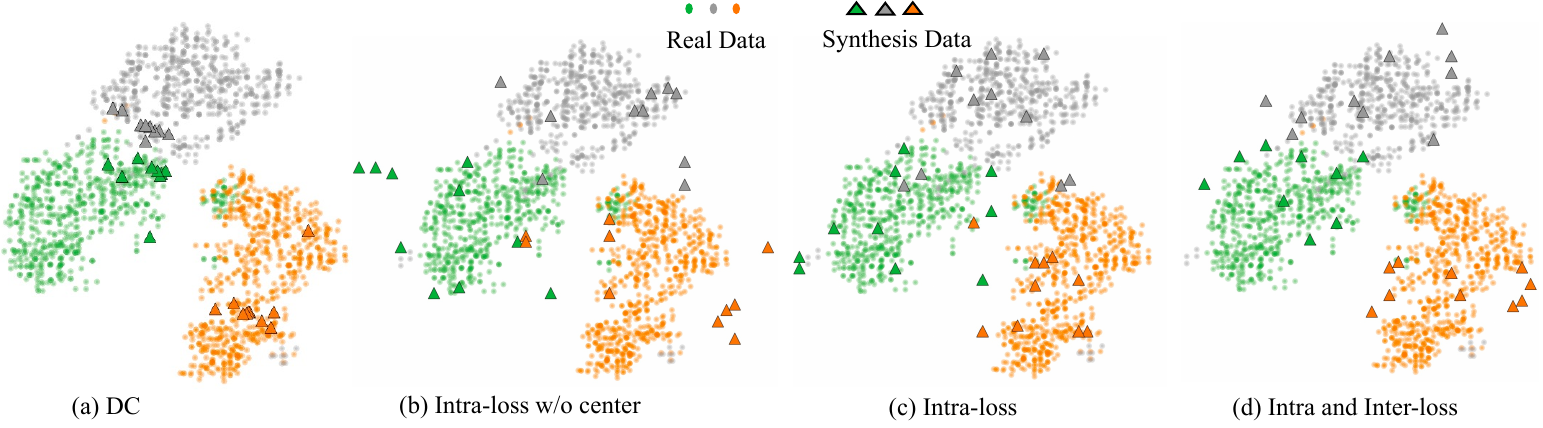}

\caption{ Visualization for distribution of synthetic samples by DC \cite{dc} and our method with different components of losses.}

\label{fig:distribution}
\end{center}
\vspace{-0.5cm}
\end{figure*}




\noindent\textbf{Loss designs.} 
We evaluate the different loss designs as shown in Tab.\ref{loss}. All losses are equipped with feature matching. Compared with the realistic real/fake loss($l_{real}$) in GAN, our distribution real/fake loss with classification($l_{feature}$) achieves the $1.5\%$, $1.9\%$, $2.7\%$ improvement  on three datasets, respectively. This indicates that acquiring realism may lead to less informative samples. When intra-class diversity loss is added, the test accuracy is improved by  $0.2\%$ and $0.3\%$ on CIFAR-10/100 but dropped by $0.2\%$ on ImageNet-sub. This is because that classes in this dataset are not completely mutually exclusive and the margins of each class are not clear. Our intra class diversity loss may cause synthetic samples to invade the margins of other classes. With inter-class discrimination loss, the performance gain-$0.7\%, 1.7\%, 1.9\%$ is expected since the loss makes the  samples of  different classes more discriminative. With the cooperation of inter-class and intra-class loss, our methods get the best results. Because synthesis images in each class are diverse, which have strong representative capabilities, and  features of the different classes have a more clear margin at the same time.

\vspace{-0.2cm}

\subsection{Visualizations}
In this subsection, we visualize the synthetic images as well as their distribution from our method and CAFE \cite{cafe} in Fig.\ref{vis},\ref{fig:distribution}.  As shown in Fig.\ref{vis}, some classes have very similar images synthesized by CAFE \cite{cafe}. On the contrary, our synthetic images are more diverse on account of inter and intra-class loss. Moreover, as shown in Fig.\ref{fig:distribution}, our distribution of synthesis images is more diverse than DC \cite{dc}. Specifically, Fig.\ref{fig:distribution} (b) shows the distribution with intra-class diversity loss but without class center constraint. The distribution of samples with different classes is crossed because the samples are spread out too thinly. With the class center as a constraint (Fig.\ref{fig:distribution} (c)), the samples are distributed diversely and also are aligned within a meaningful class space. Moreover, equipped with inter-class discrimination loss as an addition (Fig.\ref{fig:distribution} (d)), the samples in different classes are more discriminative and have a more clear class margin. These visualizations  prove the effectiveness of our intra-class diversity and inter-class discrimination loss. 
\label{visualization}

\vspace{-0.04cm}
\section{Conclusions}
In this paper, we explore a new condensed format, a generative model for dataset condensation.  The dataset is condensed  into a generative model rather than pixels.  Such a new format brings feasibility of condensation on ImageNet-1K with diverse classes.  Equipped with intra-class  and  inter-class loss, our condensed format also achieves the state-of-the-art performance on popular benchmarks.


\vspace{0.8cm}

{\small
\bibliographystyle{ieee_fullname}
\bibliography{main}
}

\end{document}